# Capacitance Resistance Model and Recurrent Neural Network for Well Connectivity Estimation – A Comparison Study

Deepthi Sen
Texas A&M University

## 1. Introduction

In this report, two commonly used data-driven models for predicting well production under a waterflood setting – the capacitance resistance model (CRM) and recurrent neural networks (RNN) are compared. Both models are completely data-driven and are intended to learn the reservoir behavior during a waterflood from historical data. The python implementation of the CRM model used in this report is available from the associated GitHub repository[1].

## 2. Capacitance Resistance Model

The capacitance resistance model for inferring interval connectivity was initially developed at UT Austin by Yousef et al (2006). This draws upon the analogy between a reservoir system under pseudo-steady state with a resistance-capacitance (RC) circuit. The inputs to the CRM are the injection rates applied at the injectors and the outputs are the production rates at the producer wells. The production rates are computed as a function of the injection rates and several parameters that may be related to the properties of the reservoir system.

### 2.1. Theory

For the pseudo-steady state condition, the rate of decrease in pressure $p$ at any point in the drainage volume at any time is a constant given by

$$\frac{\partial p}{\partial t} = \frac{\partial \bar{p}}{\partial t} = const = -\frac{q_{net}}{c_t V_p} \quad (1)$$

where
- $c_t$ : Total compressibility
- $V_p$ : Pore volume in drainage
- $\bar{p}$ : Average pressure in drainage volume
- $q_{net}$ : Net flow out of the drainage volume

Then we have

$$c_t V_p \left(\frac{d\bar{p}}{dt}\right) = i - q \quad (2)$$

- $i$ : Injection rate
- $q$ : Production rate

Assuming a linear productivity model,

$$q = J(\bar{p} - p_{wf}) \quad (3)$$

where $J$ is the productivity index that is assumed to be constant,

---

[1] https://github.com/deepthisen/CapacitanceResistanceModel

$$\frac{dq}{dt} = J\left(\frac{d\bar{p}}{dt} - \frac{dp_{wf}}{dt}\right) \quad (4)$$

$$\frac{d\bar{p}}{dt} = \frac{1}{J}\frac{dq}{dt} + \frac{dp_{wf}}{dt} \quad (5)$$

Substituting the expression for $\frac{d\bar{p}}{dt}$ into eqn (2)

$$c_t V_p \left(\frac{1}{J}\frac{dq}{dt} + \frac{dp_{wf}}{dt}\right) = i - q \quad (6)$$

Defining $\tau = \frac{c_t V_p}{J}$ and substituting into eqn (6)

$$\tau \frac{dq}{dt} + \tau J \frac{dp_{wf}}{dt} = i - q \quad (7)$$

Rearranging and integrating over time $t = t_0$ to $t = t$:

$$q(t) = q(t_0)e^{-\frac{t-t_0}{\tau}} + \frac{e^{-\frac{t}{\tau}}}{\tau}\int_{\xi=t_0}^{\xi=t} e^{\frac{\xi}{\tau}} i(\xi) d\xi \quad (8)$$

$$+ J\left[p_{wf}(t_0)e^{-\frac{t-t_0}{\tau}} - p_{wf}(t) + \frac{e^{-\frac{t}{\tau}}}{\tau}\int_{\xi=t_0}^{\xi=t} e^{\frac{\xi}{\tau}} p_{wf}(\xi) d\xi\right]$$

Extending this to multiple wells, we may have different variations of CRM which are as follows:

### 2.1.1. CRM Tank (CRMT)

Here the entire reservoir is considered as a tank system and there are but one injector and one producer. Hence, we are only concerned with the injection and net production from the system. The general formula for CRM-T is given by Eqn (8). However, in order to adapt the model for discrete samples of historical data, we start with the particular solution to Eqn (7) written as

$$q(t) = q(t_0)e^{-\frac{t-t_0}{\tau}} + \frac{e^{-\frac{t}{\tau}}}{\tau}\int_{\xi=t_0}^{\xi=t} e^{\frac{\xi}{\tau}} i(\xi) d\xi - \frac{e^{-\frac{t}{\tau}}}{\tau}\int_{\xi=t_0}^{\xi=t} Je^{\frac{\xi}{\tau}} \frac{dp_{wf}}{d\xi} d\xi \quad (9)$$

Then we apply integration by parts to the second term to obtain

$$q(t) = q(t_0)e^{-\frac{t-t_0}{\tau}} + \left[i(t) - e^{-\frac{t-t_0}{\tau}} i(t_0)\right] - e^{-\frac{t}{\tau}} \int_{\xi=t_0}^{\xi=t} e^{\frac{\xi}{\tau}} \left(\frac{di(\xi)}{d\xi} + J\frac{dp_{wf}(\xi)}{d\xi}\right) d\xi \quad (10)$$

When applied to a field with several wells, this has an added disadvantage that the individual variations in producer bottomhole pressures (BHP) cannot be accounted for. Then the $p_{wf}$ terms are to be eliminated.

Doing so, this model contains at most 3 parameters to be fitted: $q(t_0), \tau$ and $f_F$, where the latter accounts for partial support of net injection to the well group in question.

Now we can assume various profiles of injection rates and BHP control rates. Sayarpour (2008) considers two cases:
- Injection rates are constant from $t_0$ to $t$ and BHP profile changes linearly from $t_0$ to $t$. In this case $\frac{di(\xi)}{d\xi} = 0$ and $\frac{dp_{wf}(\xi)}{d\xi} = \frac{\Delta p_{wf}}{\Delta t}$ where $\Delta t = t - t_0$

- Both injection rates and BHP profile changes linearly from $t_0$ to $t$. In this case $\frac{di(\xi)}{d\xi} = \frac{\Delta i}{\Delta t}$ and $\frac{dp_{wf}(\xi)}{d\xi} = \frac{\Delta p_{wf}}{\Delta t}$ where $\Delta i = i(t) - i(t_0)$ and $\Delta t = t - t_0$

### 2.1.2. CRM Producer (CRMP)

This model considers a field with $N_{pro}$ producers and $N_{inj}$ injectors. The control volume for this model is the drainage volume of each producer. Hence, we have $N_{pro}$ time-constant parameters: $\tau_j$ for each producer, $1 \leq j \leq N_{pro}$. Furthermore, only a fraction of the injected volume at, say the $i^{th}$ injector flows to the $j^{th}$ producer. This gives rise to yet another parameter, the gain $f_{ij}$, which is closely related to the flux allocation from injector. Starting with Eqn. 10 and extending it to multi-well case, we get

$$q_j(t) = q_j(t_0)e^{-\frac{t-t_0}{\tau_j}} + \sum_{i=1}^{N_{inj}} f_{ij}\left[i(t) - e^{-\frac{t-t_0}{\tau_j}} i(t_0)\right] \tag{11}$$
$$- e^{-\frac{t}{\tau_j}} \int_{\xi=t_0}^{\xi=t} e^{\frac{\xi}{\tau_j}} \left(\sum_{i=1}^{N_{inj}} f_{ij}\frac{di(\xi)}{d\xi} + J_j \frac{dp_{wf_j}(\xi)}{d\xi}\right) d\xi$$

As with CRMT, one can impose various assumptions on the injection and BHP profiles. Going further, we impose the condition where injection rates are constant from $t_0$ to $t$ and BHP varies linearly in the same time. Then we obtain the recursive expression, given the assumption that injection rate remains constant and BHP varies linearly from $t_{n-1}$ to $t_n$.

$$q_j(t_n) = q_j(t_{n-1})e^{-\frac{\Delta t_n}{\tau_j}} + \left(1 - e^{-\frac{\Delta t_n}{\tau_j}}\right)\left[\sum_{i=1}^{N_{inj}} f_{ij}I_i(t_n) - J_j\tau_j \frac{\Delta p_{wf_j}}{\Delta t_n}\right] \tag{12}$$

It is to be noted that Eqn (12) is what is implemented in the attached python module. However, this expression can further be expanded in terms of $q(t_0)$ as

$$q_j(t_n) = q_j(t_0)e^{-\frac{t_n-t_0}{\tau_j}} + \sum_{k=1}^{n}\left\{e^{-\frac{t_n-t_k}{\tau_j}}\left(1 - e^{-\frac{\Delta t_k}{\tau_j}}\right)\left[\sum_{i=1}^{N_{inj}} f_{ij}I_i(t_k) - J_j\tau_j \frac{\Delta p_{wf_j}^{(k)}}{\Delta t_k}\right]\right\} \tag{13}$$

### 2.1.3. CRMIP (Injector-Producer)

The control volume is further divided into injector-producer bundles in this version of CRM. This leads to more number of $\tau$ (specifically $N_{inj} \times N_{pro}$), bringing up the number of trainable parameters to $4 \times (N_{inj} \times N_{pro})$. Even though this is not covered in detail in this report, there is not much conceptual difference from CRMP, except for more granularity in defining the control volume, hence more model complexity. The general discretized formula for a case where injection rates remain constant and BHP changes linearly across time steps is given below.

$$q_j(t_n) = \sum_{i=1}^{N_{inj}} q_{ij}(t_n) \tag{14}$$
$$= \sum_{i=1}^{N_{inj}} q_{ij}(t_0)e^{-\frac{t_n-t_0}{\tau_{ij}}} + \sum_{i=1}^{N_{inj}}\sum_{k=1}^{n}\left\{\left[f_{ij}I_i(t_k) - J_{ij}\tau_{ij}\frac{\Delta p_{wf_j}^{(k)}}{\Delta t_k}\right]e^{-\frac{t_n-t_k}{\tau_j}}\left(1 - e^{-\frac{\Delta t_k}{\tau_j}}\right)\right\}$$

The production from the $j^{th}$ producer is obtained by summing over individual contributions of each injector-producer bundle ending at the producer in question.

## 2.2. Implementation

The CRM can be easily implemented in an iterative manner using a recursive expression such as Eqn (12). In this report, CRMP with constant injection rates and linearly varying BHP across each timestep was implemented into a python module[2]. The basic parts of the module are explained here

a. Initialization: A CRMP object is created by accepting a list of parameters that include $\tau, f, q(t_0)$ and $J$ (if we need to consider pressure, which is given by the include_press parameter). These are set as attributes to the object.

b. Prim_prod, inject_term, bhp_term calculates the respective terms in Eqn (12) and assigns these are object attributes at any given timestep.

c. Prod_pred: computes the predicted time series given a series of inputs. Depending whether the model is in training or not, it also computes the gradients of the loss function with respect to the parameters at each timestep.

d. Compute_grad_tau, Compute_grad_lambda, Compute_grad_q0, Compute_grad_J computes the respecting gradients. These functions are called by prod_pred while in training mode. The gradients computed are as follows:

e. Compute_loss: Calculates a scaled version of the mean squared error, given an observed production timeseries.

f. Obj_func_fit and Jac_func_fit are used to compute the loss and gradients since these are fed into the SLSQP method called during fitting

g. Fit_model takes in a list of parameters as initial guess and runs the SLSQP algorithm with non-negativity bounds on all parameters and addition constraints such as making sure that $\sum_{i=1}^{N_{inj}} f_{ij} \leq 1$. For this report, I have set this as an equality constraint, since we are working with voidage replacement ratio = 1 (There is no unaccounted influx or outflux).

## 3. Recurrent Neural Networks

RNNs have been shown to offer excellent performance in time-series prediction problems such as stock market prediction, sentence completion etc. These may also be used in reservoir engineering application such as developing a fully data-driven model of connectivity and production prediction.

RNNs have the added advantage of ease of implementation using libraries such as keras/tensorflow where training and hyperparameter tuning may be achieved easily.

---

[2] https://github.com/deepthisen/CapacitanceResistanceModel

## 3.1. Theory

The basic RNN consisted of a neural network that computes the output at any timestep given the input at that timestep and the output from the previous timestep. This is illustrated in Figure 1. Therefore,

$$y_t = \sigma(Ax_t + By_{t-1} + b) \tag{15}$$
$$y_{t-1} = \sigma(Ax_{t-1} + By_{t-2} + b)$$
$$\vdots$$
$$y_1 = \sigma(Ax_1 + By_0 + b)$$

Here, $A, B$ represent the kernel and recurrent weights respectively and $b$ represents the bias and $\sigma$ represents the activation function. In this application, the activation function has been set to linear. Additionally, ignoring the bias, at any timestep:

$$y_t = Ax_t + B(Ax_{t-1} + B(\ldots + By_0)) \tag{16a}$$
$$y_t = A(x_t + Bx_{t-1} + B^2 x_{t-2} + \cdots + B^{t-1}x_1) + B^t y_0 \tag{16b}$$

However, we specify a time-window (TS) over which this kind of calculation happens:

$$y_t = A(x_t + Bx_{t-1} + B^2 x_{t-2} + \cdots + B^{t-1}x_{t-TS}) + B^t y_{t-TS-1} \tag{17}$$

In this application, we use '**stateless RNNs**', where for calculation at each time step, $y_{t-TS-1}$ is set to zero. Stateful RNNs set the $y_{t-TS-1}$ at the actual predicted value of this timestep. However, the training process becomes slightly more involved with the need to choose batchsize carefully to prevent overlap between successive batches.

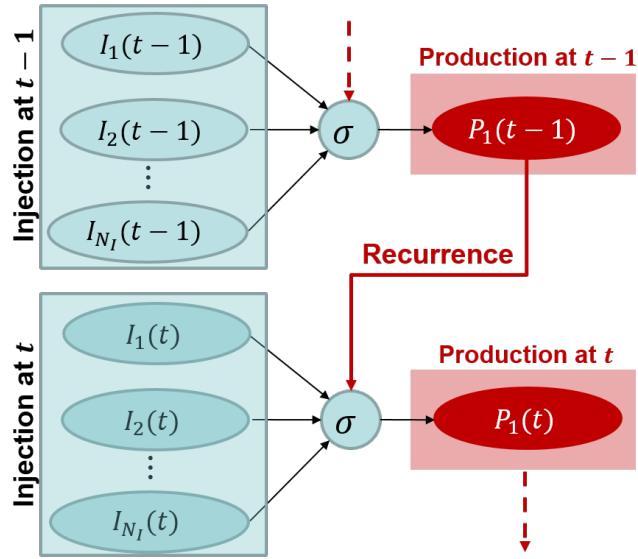

Figure 1 Schematic illustration of the recurrent neural network implemented

## 3.2. Implementation

A custom RNN layer was implemented with additional constraints on
- the $A$ matrix to impose location constraints in case of large field applications. This has not been imposed on the problems covered in this report.

- the *B* matrix to limit the influence of producers on each other. This may be relaxed if needed.

As mentioned before, the activation function used is linear and bias has been set to zero.

## 4. Case Studies
### 4.1. Case Study 1: Streak Case
#### 4.1.1. Description

This synthetic case with 5 injectors and 4 producers has been widely studied in CRM literature. It is a homogenous field with two high permeability streaks that connect two pairs of injector-producer. The streak case permeability field is shown in Figure 2 and the prescribed injection rates for the five injectors are shown in Figure 3.

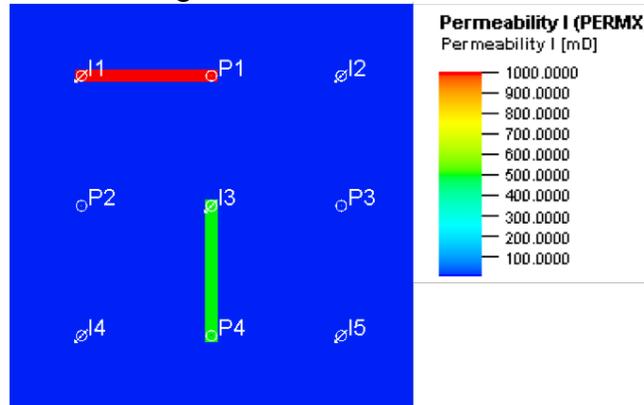

Figure 2 Permeability field of Synthetic Case 1 (Streak Case)

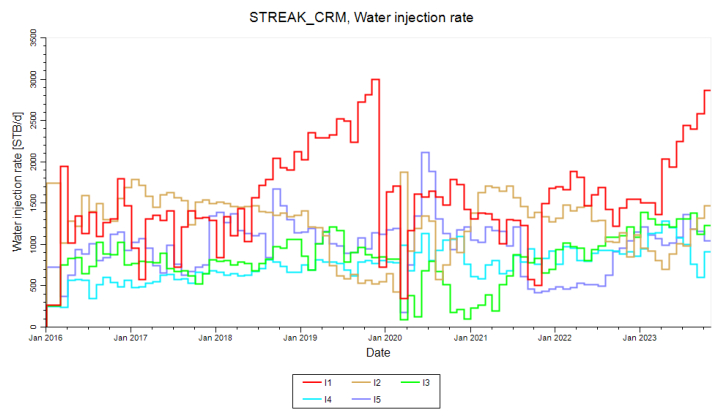

Figure 3 Injection rates prescribed in the Streak Case. These were obtained from Sayarpour (2008).

#### 4.1.2. Results and Discussion

It is seen from Figure 4 that the prediction accuracies for the producers P1 and P4, which are connected to the injectors via the high-perm streaks, are very high for both RNN and CRMP. However, there is significant degradation in performance at test time for P2 and P3, especially in CRMP. This may be attributed to the non-linearity in the response of these producers to the input signals.

Insight on the connectivity of the field can be gained from both the fitted CRMP parameters as well as the trained RNN weights.

Table 4.1 and Table 4.2 show the fitted CRM parameters. Theoretically, $\tau$ represents the relative sizes of the drainage volumes associated with each producer. However, it is seen that the objective function is highly non-convex in $\tau$ and the converged values of $\tau$ are highly dependent on the initial condition. This effect is greatly amplified in the optimization for $q(t_0)$ where the gradients with respect to these are extremely small and there is not much optimization taking place by varying $q(t_0)$.

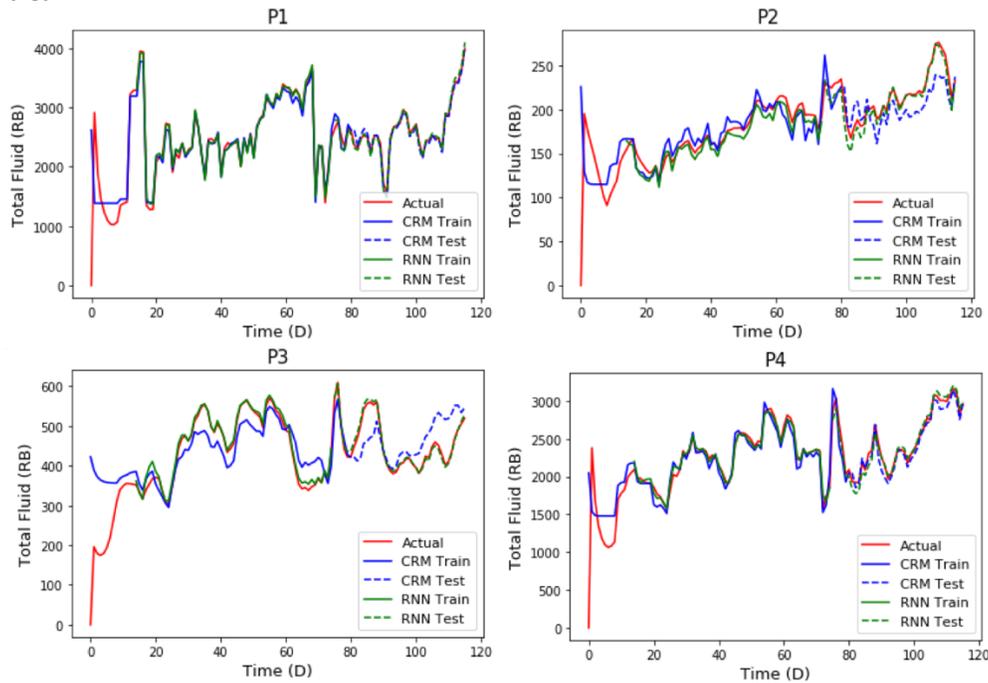

Figure 4 Performance of CRMP and RNN on the streak case of Fig .The red plot represents the actual total reservoir volumes production from each producer. The blue and green continuous line represents the training time predict of the CRMP and RNN respectively, whereas the corresponding broken lines represent the test time prediction.

Table 4.1 CRMP fitted parameters for streak case: $\tau$ and $q(t_0)$

|  | P1 | P2 | P3 | P4 |
|---|---|---|---|---|
| $\tau$ | 0.1 | 0.5 | 1.7 | 0.6 |
| $q(t_0)$ | 0.0 | 0.25 | 0.02 | 0.01 |

Table 4.2 CRMP Fitted gain matrix for the streak case.

|  | P1 | P2 | P3 | P4 |
|---|---|---|---|---|
| I1 | 0.95 | 0.02 | 0.00 | 0.03 |
| I2 | 0.51 | 0.03 | 0.12 | 0.29 |
| I3 | 0.03 | 0.00 | 0.10 | 0.87 |
| I4 | 0.18 | 0.12 | 0.07 | 0.63 |
| I5 | 0.16 | 0.02 | 0.16 | 0.66 |

However, the most valuable information is contained in the gain matrix $f$ (Table 4.2). There exists a direct correlation between this and the actual rate allocation factor from injector to producer (in fact, by definition). An illustration of the gain matrix is provided in Figure 5(a), where it is clear how the higher values for $f_{ij}$ for a particular injector-producer pair corresponds to those with high physical connectivity in terms of volumetric flux (as observed from the number of streamlines between a well-pair as in Figure 5(c)).

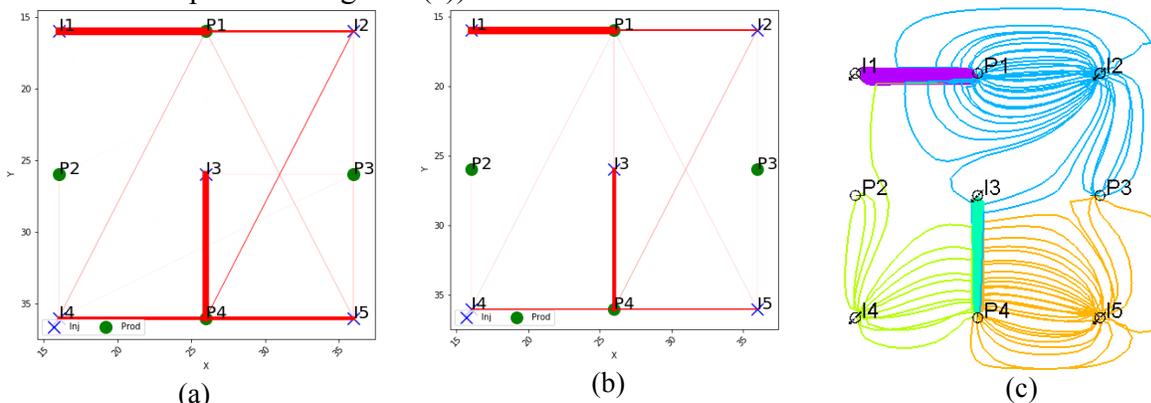

Figure 5 Comparison of connectivities inferred from (a) CRMP (b) RNN and (c) Flow simulation via streamlines

In the case of RNN, the connection between the injectors and producers are signified by the kernel weight matrix $A$. These are shown in Table 4.3 and schematically illustrated in Figure 5(b). The concurrence between the simulation-based connectivity and the A matrix is clear from Figure 5. The recurrence weight matrix $B$ represents the influence that the previous value of the production has over the current production (Table 4.4).

Table 4.3 Trained kernel weights A for RNN model for streak case

|    | P1   | P2   | P3   | P4   |
|----|------|------|------|------|
| I1 | 0.95 | 0.01 | 0.00 | 0.00 |
| I2 | 0.45 | 0.01 | 0.09 | 0.18 |
| I3 | 0.15 | 0.01 | 0.00 | 0.68 |
| I4 | 0.11 | 0.09 | 0.00 | 0.39 |
| I5 | 0.11 | 0.00 | 0.09 | 0.40 |

Table 4.4 Trained recurrence weights B for RNN model for streak case

|    | P1   | P2   | P3   | P4   |
|----|------|------|------|------|
| P1 | 0.04 | 0    | 0    | 0    |
| P2 | 0    | 0.39 | 0    | 0    |
| P3 | 0    | 0    | 0.50 | 0    |
| P4 | 0    | 0    | 0    | 0.35 |

## 4.2. Case Study 2: Non-Streak Case

In order to further test the efficacy of using CRMP and RNN when the responses are highly nonlinear functions of the input, I removed the streak cases and generated a fully homogenous (1 mD permeability) model with the same injectors and producers (Figure 6), with the effect that the nonlinearity in the responses at all producers is more noticeable (Figure 7).

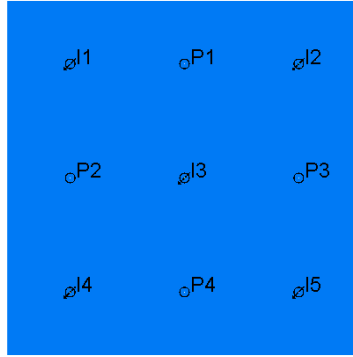

Figure 6 Homogenous permeability and porosity field with the same well configuration as the streak case.

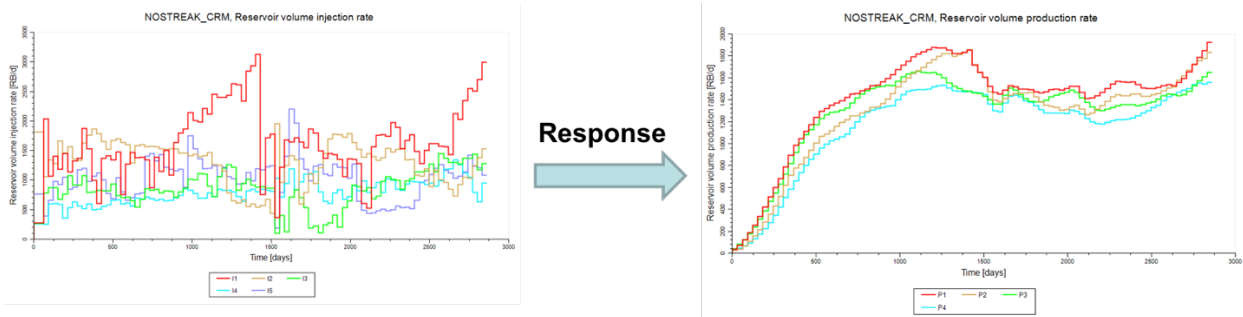

Figure 7 The injection profile is the same as the streak case. However, the responses from the producers are high linear.

The results obtained by fitting a CRMP and an RNN are shown in Figure 8. The inability of CRM at capturing nonlinearity is more apparent at all wells in this case. However, it is again seen that the RNN does a better job, both at training and test time. This may be attributable to the window-based way in which the stateless RNN is trained. Hence the effect of fitting the early part of the production profile on the prediction of the latter portion (at test time), is reduced.

Table 4.1 and Table 4.2 show the fitted CRMP parameters. From Figure 9(a), the CRMP seems to have picked up spurious connectivity trends, such the increased connection between I1- P2 and I5- P3, despite this being a completely homogeneous case. However, the injector-producer connectivities obtained from the RNN (Figure 9(b)) is more representative of the homogeneity since all connections are almost equally weak, as seen in Table 4.5.

Furthermore, the recurrence weights $B$ for the RNN (Table 4.6) also clearly shows higher values in comparison with those for the streak case (Table 4.4)

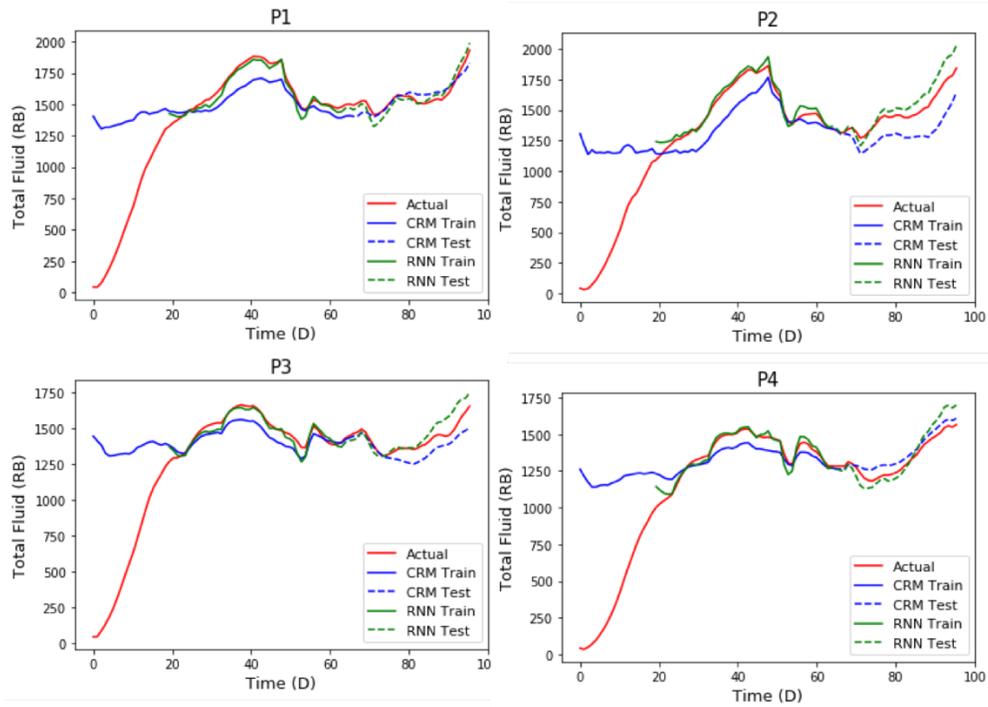

Figure 8 Performance of CRMP and RNN on the non-streak case of Figure 6. The red plot represents the actual total reservoir volumes production from each producer. The blue and green continuous line represents the training time predict of the CRMP and RNN respectively, whereas the corresponding broken lines represent the test time prediction.

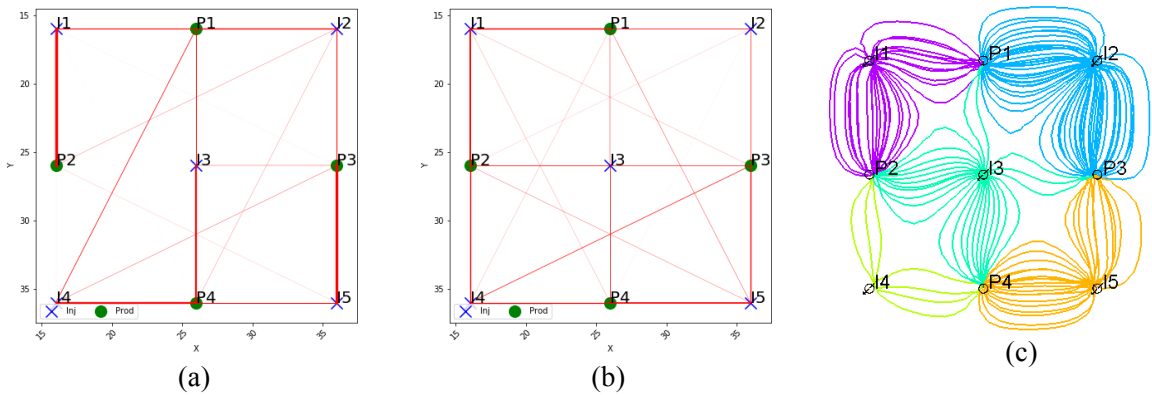

(a)      (b)      (c)

Figure 9 Comparison of connectivities for the non-streak case inferred from (a) CRMP (b) RNN and (c) Flow simulation via streamlines

Table 4.5 Trained kernel weights A for RNN model for non-streak case

|    | P1   | P2   | P3   | P4   |
|----|------|------|------|------|
| I1 | 0.09 | 0.10 | 0.03 | 0.03 |
| I2 | 0.06 | 0.02 | 0.06 | 0.01 |
| I3 | 0.05 | 0.06 | 0.06 | 0.06 |
| I4 | 0.03 | 0.08 | 0.06 | 0.08 |
| I5 | 0.04 | 0.04 | 0.08 | 0.09 |

Table 4.6 The recurrence weight B for the RNN model for the non-streak case

|    | P1   | P2   | P3   | P4   |
|----|------|------|------|------|
| P1 | 0.79 | 0    | 0    | 0    |
| P2 | 0    | 0.76 | 0    | 0    |
| P3 | 0    | 0    | 0.79 | 0    |
| P4 | 0    | 0    | 0    | 0.79 |

## 4.3. Computational Efficiency

Despite better performance with respect to non-linearity, the RNN model built on Keras/Tensorflow is slower during training due to the need to run several epochs of fitting for the weights to converge. However, the number of epochs at training is tunable and procedures like early-stopping etc. may be useful in bringing down the required number of epochs. Nevertheless, this is unlikely to be comparable to the time required to fit a CRM using an optimization method such as SLSQP, which is lesser by an order of magnitude, as seen in Table 4.7

Table 4.7 A comparison of CPU times for CRMP vs. RNN during training for the two synthetic cases

| Model | CPU Time(s) |
|---|---|
| CRM – Streak | 0.2 |
| RNN – Streak (500 epoch) | 5.6 |
| CRM – No Streak | 0.1 |
| RNN – No Streak (500 epoch) | 5.5 |

At test time, both CRMP and RNN are quite comparable and both are faster than running a conventional simulator by several orders of magnitude, as is seen in Table 4.8.

Table 4.8 A comparison of CPU times for CRMP, RNN vs. commercial simulator for the two synthetic cases

| Model | CPU Time(s) |
|---|---|
| CRM – Streak | 0.003 |
| RNN – Streak (500 epoch) | 0.006 |
| *Eclipse – Streak* | *2.3* |
| CRM – No Streak | 0.004 |
| RNN – No Streak (500 epoch) | 0.006 |
| *Eclipse – No Streak* | *4.7* |

## 5. Future Work

The synthetic case-2 (non-streak case) illustrates a situation where the CRMP fails to capture the reservoir dynamics perfectly. The next logical improvement to the current work will be to implement a model of the next level of complexity: the CRMIP. This has been left for future work.

Furthermore, one may incorporate a better activation function for the RNN model, that does a better job at accounting for non-linear dynamics. Other ways of trying to incorporate nonlinearity effectively includes the use of stacked RNNs or perhaps a better RNN variant such as LSTM.